\documentclass[letterpaper, 10 pt, conference]{ieeeconf}  

\IEEEoverridecommandlockouts                              

\usepackage{graphics} 
\usepackage{graphicx}
\usepackage{subcaption}
\usepackage{epsfig} 
\usepackage{mathptmx} 
\usepackage{times} 
\usepackage{amsmath} 
\usepackage{amssymb}  
\usepackage[dvipsnames]{xcolor}
\usepackage{bbm}
\usepackage{hyperref}
\usepackage{float}
\usepackage{siunitx}
\usepackage{cite}
\usepackage{tikz}
\newcommand{\rowlabel}[1]{#1\vphantom{pg}}
\newcommand{\edit}[1]{#1}
\usepackage{filecontents}

\DeclareMathAlphabet{\mathbbold}{U}{bbold}{m}{n}

\title{\LARGE \bf
GeCCo - a Generalist Contact-Conditioned Policy for Loco-Manipulation Skills on Legged Robots}

\author{Vassil Atanassov$^{\dagger,1}$, Wanming Yu$^{1}$, Siddhant Gangapurwala$^{2}$, James Wilson$^{3}$, Ioannis Havoutis$^{1}$%
\thanks{%
$^{1}$Machine Learning and Articulated Robots Group, Oxford Robotics Institute, University of Oxford,
$^{2}$Sony AI,
$^{3}$Dyson Institute of Engineering and Technology,
$^{\dagger}$Corresponding author: {\tt vassilatanassov@robots.ox.ac.uk}%
}}

\begin{document}
\bstctlcite{IEEEexample:BSTcontrol} 

\maketitle
\thispagestyle{empty}
\pagestyle{empty}

\begin{abstract}    
Most modern approaches to quadruped locomotion focus on using Deep Reinforcement Learning (DRL) to learn policies from scratch, in an end-to-end manner. Such methods often fail to scale, as every new problem or application requires time-consuming and iterative reward definition and tuning.
We present Generalist Contact-Conditioned Policy (GeCCo) --- a low-level policy trained with Deep Reinforcement Learning that is capable of tracking arbitrary contact points with a quadruped robot. 
We shift from task-specific end-to-end learning to a modular hierarchy in which a single planner-agnostic, contact-conditioned tracking policy serves as a robust interface between planning and control. By maintaining stable contact execution under dynamic uncertainty and planner mismatch, the policy enables high-level planners (e.g., handcrafted, learned, or optimization-based) to be composed and swapped without retraining.
We demonstrate the scalability and robustness of our method by evaluating on a wide range of locomotion and manipulation tasks in a common framework and under a single generalist policy. These include a variety of gaits, traversing complex terrains (e.g., stairs and slopes) as well as previously unseen stepping-stones and narrow beams, and interacting with objects (e.g., pressing a button, pushing a barrel). Our framework acquires new behaviors more efficiently, simply by combining a task-specific high-level contact planner and the pre-trained generalist policy. Videos: https://vassil-atn.github.io/gecco.github.io
\end{abstract}

\section{Introduction}
From dense foliage and forests, to rocky mountains and caves, the morphology of legged robots provides a solution with better agility and utility in all the terrains surrounding us. But realizing this potential requires grappling with the complexity of control: adapting gaits, transitions, and motion strategies to the irregular demands of varied terrains. In the last few years, Deep Reinforcement Learning (DRL) has made legged locomotion controllers significantly more robust, computationally efficient, and capable of ever-more difficult tasks. DRL approaches can now not only traverse natural environments, but actively interact with the world.

While state-of-the-art controllers achieve impressive results, many of them aim to solve just one specific task --- for example to jump \cite{rudin_cat-like_2022,atanassov_curriculum-based_2025}, open doors \cite{schwarke_curiosity-driven_2023} or manipulate objects \cite{jeon_learning_2024,arm_pedipulate_2024} to name a few. This typically involves defining the task specifications and reward functions, and fine-tuning their (likely interconnected and co-dependent) scales. As a result, these methods only show a single well-tuned behavior and often do not generalize across tasks. If we want the robot to exhibit new behaviors, or succeed in new unseen environments, we would need to retrain the policy from scratch. This can result in a significant computational burden, especially as many state-of-the-art approaches rely on first training a teacher policy, and then distilling it into a student network \cite{chen_learning_2020,miki_learning_2022}. Furthermore, these methods often use a command-space designed for the specific task in mind --- for example, this is commonly a linear and angular velocity target \cite{miki_learning_2022}, goal-position for the robot to reach \cite{rudin_advanced_2022,atanassov_constrained_2024, hoeller_anymal_2023}, desired jumping targets \cite{atanassov_curriculum-based_2025}, or end-effector targets \cite{arm_pedipulate_2024}. Unfortunately, this means that we cannot easily distill all of these controllers into a single policy, unless we predict beforehand and pad for all possible commands, masking out the unnecessary ones for each individual task. 
\begin{figure}
    \centering
    \includegraphics[width=\columnwidth]{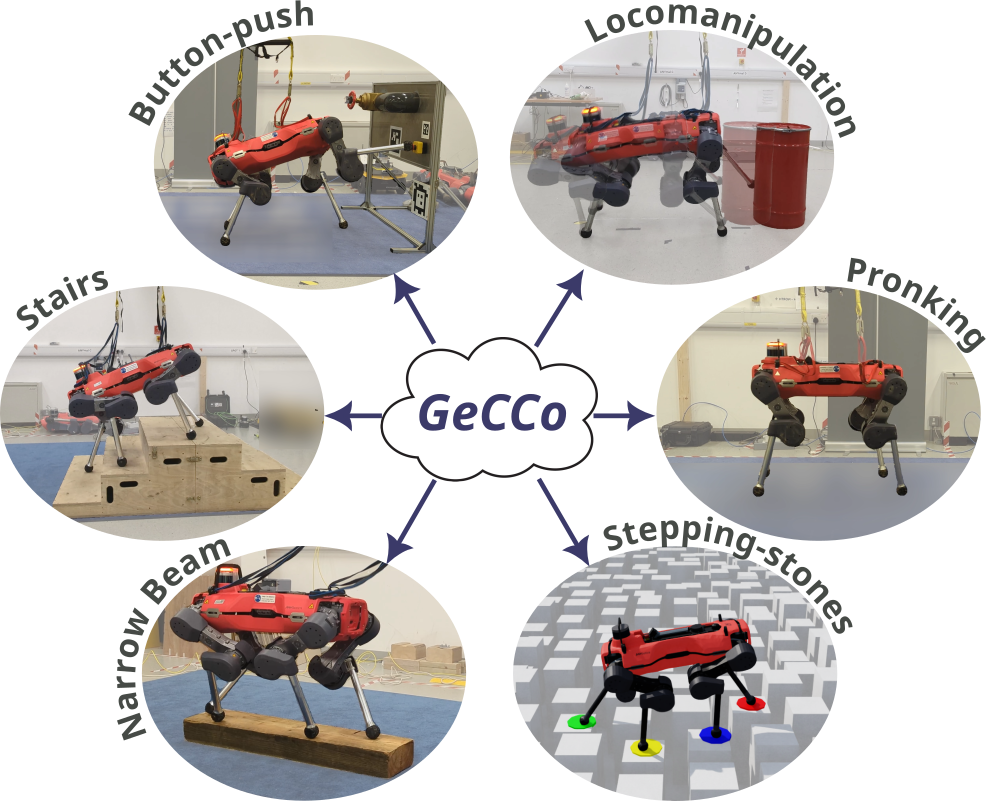}
    \caption{A selection of diverse and distinct robot behaviors resulting from a single policy trained with GeCCo.}
    \vspace{-4pt}
    \label{fig:teaser}
\end{figure}

\begin{figure*}
    \centering
    \includegraphics[width=\textwidth]{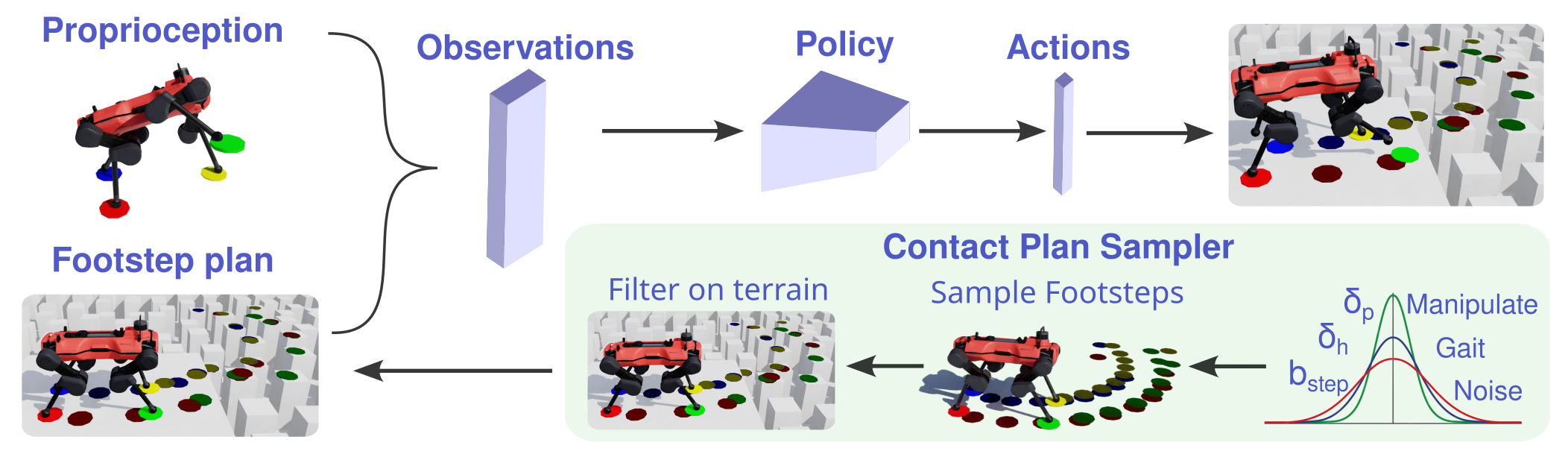}
    \caption{General scheme of the GeCCo framework. The planner samples a desired contact plan, based on several distributions. The policy then receives the observations, consisting of the proprioception and the current footstep target, and produces desired joint angles as the actions.}
    \vspace{-4pt}
    \label{fig:diagram}
\end{figure*}

Current state-of-the-art approaches are limited in their generality, as these mostly use the aforementioned command space structure. To overcome this, we propose a generalist contact-conditioned framework, named GeCCo hereafter, \edit{which learns to track desired contact locations and durations for each foot.} By decoupling the problem into planning and control, we can train a controller that generalizes better than an end-to-end policy, effectively reframing the task as an in-distribution generalization problem. \edit{In this work, we focus specifically on the \textbf{tracking} problem, and use a simple planner for both training and inference. Despite the planner not taking into account kinematic and dynamic constraints, GeCCo is capable of learning a robust policy that can accurately track the feet locations, even in dynamic, real-world environments.}
We demonstrate GeCCo's utility on a wide range of tasks on the quadruped ANYmal robot --- such as walking, jumping, manipulating objects, and traversing complex terrains (see Fig. \ref{fig:teaser}), and our framework can easily scale to robots with fewer or more contact points. While contact-tracking controllers have been proposed in prior work \cite{jenelten_dtc_2023,zhang_wococo_2024,kim_high-speed_2025}, to the best of our knowledge, GeCCo is the first one to do this in a general way, across a variety of tasks, locomotive gaits and (non-prehensile) loco-manipulation skills under a single policy. Thanks to our sampling strategy, GeCCo can be trained quickly and efficiently in about half a day on a single GPU.

Our contributions are as follows:
\setlength{\topsep}{2pt}
\begin{itemize}
    \setlength{\itemsep}{1pt}
    \setlength{\parskip}{0pt}
    \setlength{\parsep}{0pt}
    \item We present a novel generalist contact-conditioned (GeCCo) policy for learning to track arbitrary foot plans. Using a sampling-based approach to generate desired contact locations, we can train general policies more efficiently.
    \item We demonstrate that the policy can exhibit different styles and locomotive gaits, and successfully traverse complex terrains (staircases, narrow beams, and stepping stones).
    \item We combine both locomotion and manipulation capabilities within a single policy, which enables the robot to seamlessly combine walking and interacting with objects using its feet as non-prehensile manipulators.
    \item We deploy GeCCo in the real world using a simple heuristic-based planner, showing strong robustness and adaptability in tracking the contact plan, despite the planner not considering kino-dynamic constraints.
\end{itemize}

\section{Relevant Work}

\textit{Legged locomotion:} Model-based control methods often decompose the task of locomotion into several sub-components: perception and localization modules that produce elevation maps and state estimates, high-level planners that generate optimal trajectories for the feet and the center of mass, and low-level controllers (sometimes also hierarchical) that aim to track these trajectories \cite{grandia_perceptive_2022,papatheodorou_momentum-aware_2024}. 

In the last several years, deep neural networks trained with reinforcement learning have shown incredible performance when it comes to both blind \cite{hwangbo_learning_2019, lee_learning_2020} and perceptive locomotion \cite{miki_learning_2022,gangapurwala_rloc_2022}. Most commonly, the locomotion task is framed as a base velocity-tracking or as a goal-conditioned one \cite{rudin_advanced_2022, hoeller_anymal_2023}. 
Sparse terrains (like stepping-stones), however, have remained particularly difficult for such locomotion controllers. Recent work \cite{zhang_learning_2023} has shown success by training separate specialist policies for stepping-stones and narrow beams. Other works \cite{he_attention-based_2025} have done this with a single policy with an attention encoder, but require encountering those environments during training, and have long training times (several days \cite{he_attention-based_2025}). On the other hand, our proposed GeCCo framework can handle these terrains zero-shot, without training on them.

Reinforcement learning has also been used to learn policies capable of whole-body loco-manipulation using the limbs of the robot. These approaches learn a single task using a dedicated end-to-end policy, to dribble \cite{ji_dribblebot_2023} and manipulate objects to a desired location \cite{jeon_learning_2024}. Recent work \cite{cheng_legs_2023, arm_pedipulate_2024, cheng_rambo_2025} has made this more general by learning to track desired positions using the feet of a quadruped robot. However, \cite{cheng_legs_2023, arm_pedipulate_2024} both learn a specific manipulation policy, and require a separate locomotion policy to walk normally. In contrast, GeCCo can do this naturally under the same framework, allowing a seamless combination of manipulation and locomotion, as shown in Section \ref{subsect:results_locomani}.

\textit{Contact-tracking:} Recently, several works mirrored model-based controllers by learning contact-conditioned policies \cite{jenelten_dtc_2023,zhang_wococo_2024,kim_high-speed_2025}. Similarly to MPC locomotion controllers, there is a high-level planner (or heuristically-defined plans) that computes a foothold schedule, which the locomotive policy aims to track. DTC \cite{jenelten_dtc_2023} used Trajectory Optimization (TO) to compute a footstep plan for the RL controller to track, which, however, leads to long training times (14 days). WoCoCo \cite{zhang_wococo_2024} learns from manually-defined plans, and shows that the same reward can be used to learn locomotion, manipulation, and other controllers. However, in contrast to our proposed approach, a separate policy is trained for each task in WoCoCo, and it requires exploration terms to combat the sparsity of the rewards. In \cite{kim_high-speed_2025}, a sampling-based planner is learned alongside the tracking policy, allowing both controllers to influence each other as they improve. On the other hand, to shrink the learning space and reduce computational cost, the authors only consider a single bounding gait. In GeCCo, as we sample from a large set of contact locations and timings, the policy can instead achieve a much wider range of gaits.
\section{Methodology}
\subsection{Contact-commands space} \label{subsect:contact_commands_space}
To train the GeCCo contact-tracking framework, at the beginning of each episode we sample $\{\mathcal{R}_{i}\}_{i=1}^{N_{\text{stages}}}$ for all contact points, where $N_{\text{stages}}$ is the number of contact stages.
Each $\mathcal{R}_i$ is a target region for a contact point, defined as a sphere in $\mathbb{R}^3$:  
\[
\mathcal{R}_i = \{\, p_i \in \mathbb{R}^3 \;|\; \|p_i - c_i\|_2 \leq r \,\}, \quad r = 0.1 \,\text{m},
\]  
centered at $c_i$. For a quadruped, this gives four regions per stage, one for each foot. A contact is satisfied for a foot when its position $p_i \in \mathcal{R}_i$ and the foot is in ground contact. In practice, this collapses to the support surface,  
\[
\tilde{\mathcal{R}}_i = \{\, p_i \in \mathcal{R}_i \;\mid\; (p_i)_z = z_{\text{ground}} \,\},
\]  
so contact satisfaction reduces to $p_i \in \tilde{\mathcal{R}}_i$. For loco-manipulation tasks, however, sampling valid contact points directly on an object in $\mathbb{R}^3$ is difficult. To keep the sampling tractable, we drop the ground-contact condition and only require the foot to lie within the tolerance region, i.e.\ $p_i \in \mathcal{R}_i$. \edit{We tested several values for the tolerance region $r$, and chose $0.1$\unit{m} as it strikes a good balance between accurate tracking and ease of learning.}


Each contact point has an associated contact duration, $N_{dur}$, which defines how long a foot should remain in contact before transitioning to the next stage. \edit{To allow for variable stance duration at inference,} and to accommodate gaits that require foot-specific timing, we also introduce randomized durations, allowing each foot (or different combinations thereof) to maintain its own counter and progress independently. In typical cases, we use a uniform duration across all four feet, with progression occurring only once every foot has satisfied its assigned contact time.

\textit{Sampling:} A crucial part of getting a generalist contact-tracking policy is learning to track a wide range of contact points. Using a model-based planner \cite{jenelten_dtc_2023}, or learning it alongside the policy \cite{kim_high-speed_2025} risks overfitting to that particular planner. Instead, we choose to sample a large set of possible contacts, which is computationally simpler while covering more of the state space. Inevitably, some of these contact points might not be feasible or stable. However, in our experience, this did not impede training, and helped robustify the policy, by exposing it to harder combinations of contacts. In general, we sample $N_{stages}$ deviations of the base position $\delta_p$, and heading deviations $\delta_h$. We then propagate forward the desired base position. The contact points are then sampled from a zero-mean Gaussian distribution $\mathit{N}(0,0.05\text{m})$ centered around the nominal location of each foot at the corresponding desired base position. \edit{We chose this value, as it allows for the robot to see a larger range of points, without sacrificing feasibility by sampling too far from the robot.} We refer to Table \ref{tab:sampling} for all of the sampling ranges.
%
%

To sample a wide range of behaviors, we define several skeleton gaits (trot, pronk, pace, bound, and single-step gait). These provide a structured yet compact basis that captures the principal coordination patterns observed in legged locomotion. From this foundation, we introduce variability by randomizing the contact durations across different feet. This combination allows us to generate a rich space of behaviors that extends far beyond the original skeletons, effectively balancing pre-engineered gait designs with the flexibility to explore diverse and nonstandard gaits. It is worth noting that the policy is not explicitly aware of the gait -- it only observes the desired contact points for the current step. At the start of each episode we also sample whether the robot should locomote or manipulate (or a combination of both)\footnote{We dynamically shift this probability proportionally to the mean contact stage of locomotion relative to manipulation.}.

For more complex terrains, like stairs and stepping-stones, we need to adjust the pre-planned footsteps. To make sure the steps lie on a surface, we project them to the nearest horizontal surface, while avoiding being too close to an edge. We also filter out contacts that would be too far away from the body ($\geq1$\unit{m}) to eliminate completely infeasible plans.
\begin{table}[h]
    \centering
    \small 
    \renewcommand{\arraystretch}{0.9}
    \caption{Sampling ranges for GeCCo.}
    \label{tab:sampling}
    \begin{tabular}{ll}
        Name (Unit) & Sampling Range \\
        \hline
        Base pos. dev. $\delta_p$ (m) & $\mathit{U}(0.0,0.4)$ \\
        Base head. dev. $\delta_h$ (rad) & $\mathit{U}(-\pi/6, \pi/6)$ \\
        Footstep noise (m) & $\mathit{N}(0.0,0.05)$ \\
        Footstep width $b_{step}$ (m) & $\mathit{N}(0.2,0.1)$ \\
        Contact duration (step) & $\mathit{U}(1,50)$ \\
        Independent feet  & $\mathrm{Ber}(0.1)$ \\
        Direction & $\mathrm{Ber}(0.5)$ \\
        Walk, Manipulate, Both & $\mathrm{Binomial}(0.5, 0.4, 0.1)$ \\
        Manip. range ($x,y,z$) (m) & $\mathit{U}(\pm[0.4,0.3,0.6])$ \\
        Gaits (Trot, Pace, Bound, etc.) & $\mathrm{Binomial}(0.2)$ \\
    \end{tabular}
    \vspace{-15pt}
\end{table}

\subsection{Observation and Action space}
We use an Asymmetric Actor-Critic \cite{pinto_asymmetric_2018} framework, where the critic observations contain privileged information. The actor observations, $o_{\text{actor}} \in \mathbb{R}^{81}$, consist of the base linear and angular velocities, projected gravity vector, foot contact states, foot position (base frame), contact commands error (base frame), a boolean indicating whether the contacts are satisfied (and at the previous step), joint positions, velocities, and previous actions. We also tested a variation that included the elevation around the robot as an exteroceptive observation, but it did not result in better performance (see Section \ref{subsect:results_ablation}).
The critic ($o_{\text{critic}} \in \mathbb{R}^{110}$) contains the same observations, but without added noise, as well as the foot velocities, the time at which correct contact was made, the elapsed time in the current contact stage, the desired contact duration, a flag to indicate if a given foot should be manipulating, and base height. The actions consist of the desired joint positions for the 12 joints, relative to the nominal stance configuration.
\subsection{Curriculum}
As in prior work \cite{hwangbo_learning_2019,rudin_learning_2022}, we employ a curriculum that automatically increases the difficulty of the environment as the agent learns. Completing at least 10 stages by the end of the episode increments the curriculum level, and reaching fewer than 5 decrements it. The terrain levels are scaled by the curriculum, as in \cite{rudin_learning_2022}, and so are some penalties - as described in Section \ref{subsect:rewards}.
\subsection{Rewards} \label{subsect:rewards}
\subsubsection{Task Rewards}
We use one main event-driven task reward, similar to that of \cite{zhang_wococo_2024} and \cite{xue_full-order_2024}, that consists of four terms. It is worth noting that this is a general reward and does not favor certain gaits, rather it only ensures the correct contacts are reached, as follows:


\begin{equation}
\label{eq:task_rew}
\begin{aligned}
r_{\text{task}}
={}& \gamma_{\text{rew}}\Big(
      \underbrace{n_{\text{corr}}}_{\text{Correct}}
      + \underbrace{\tfrac{50}{N_{\text{dur}}}\,
        \mathbbold{1}[\,n_{\text{corr}} = n_{\text{total}}\,]}_{\text{Completion bonus}}
    \Big) \\
-{}& \gamma_{\text{pen}}\Big(
      \underbrace{n_{\text{wrong}} - n_{\text{corr,prev}}}_{\text{Wrong}}
      + \underbrace{n_{\text{total}}\,n_{\text{broken}}}_{\text{Correct broken}}
    \Big)
\end{aligned}
\end{equation}

where $n_{\text{corr}}$, $n_{\text{wrong}}$, $n_{\text{total}}$ are the number of currently correct, incorrect, and total contacts (in our case this is four, one for each foot); $n_{\text{corr\_prev}}$ are the contacts that were correct at the previous contact stage and have not moved since, $ n_{\text{broken}}$ are the number of contacts that are no longer correct.

The first two terms reward correct contacts and give a bonus when all contacts are correct, respectively. The latter two terms penalize wrong contacts -- this can lead to large negative rewards as soon as the agent receives a new target, so we temporarily discount the penalty as long as the foot remains at the previously correct point \cite{xue_full-order_2024} -- and breaking correct contacts as we want the agent to continuously maintain these contacts. 
We then use two scales for the different terms: $\gamma_{pen}$ for the penalties, which is 0 initially, and increases by a factor of $0.25 \times \mathrm{curriculum\ level}$; and $\gamma_{rew}$ for the rewards, which is defined as follows:
\begin{equation}
\label{eq:gamma_rew}
\gamma_{\text{rew}} = \frac{1}{2}\,
  \underbrace{\bigl(1 + \exp(-100\,d_{\text{cntc}}^{2})\bigr)}_{\text{proximity}}
  \cdot
  \underbrace{\bigl(1 + \exp(-(\sigma_{\text{td}}/2)^{2})\bigr)}_{\text{variance}},
\end{equation}
where $d_{\text{cntc}}\in[0,0.1]$\unit{m} is the distance between the desired and actual contact, and $\sigma_{td}$ is the variance of the touchdown time for the feet whose contact target has changed. 
The proximity term encourages stepping closer to the center of the target. The variance term rewards moving feet in synchronicity, and helps with tracking different gaits. 
\subsubsection{Regularization Rewards}
To regularize the motion and allow it to transfer to the real hardware, we use several smoothing rewards (see Table \ref{tab:rewards}), which have been shown \cite{lee_learning_2020,rudin_learning_2022} to be important for successful sim-to-real transfer. 
To discourage dragging the feet when moving from one footstep to the next, we employ a clearance penalty as in \cite{kim_contact-implicit_2024}. 
In addition, as the task reward could encourage quick motions, after a certain curriculum level a barrier-function-styled cost \cite{zhang_constrained_2024} kicks in to penalize exceeding foot velocity above $1.75$\unit{m/s} for locomotion and $0.5$\unit{m/s} for manipulation.
\subsubsection{Terminations}
To ensure the policy is safe to deploy, and to facilitate learning, we terminate upon body contacts, and if the contact stage has not progressed within $8$\unit{s}.
\begin{table}[]
    \centering
    \small
    \renewcommand{\arraystretch}{0.9}
    \caption{Reward terms and scales. $n_j$, $n_c$, $n_f$ denote the number of joints, collision points, and feet; the smooth barrier function $bf(\cdot)$ is defined in \cite{zhang_constrained_2024}.}
    \begin{tabular}{c|c|c}
        Reward Name & Function & Scale  \\
        \hline
        Task & Eq. \ref{eq:task_rew} & 5.0 \\
        Base Height & $\text{exp}(-400 ||h - 0.6||^2)$ & 10.0 \\
        Alive Bonus & $\mathbbold{1}_{\text{alive}}$ & 10.0 \\
        Flat Orientation & $||g_{\text{proj}}||^2$& -5.0 \\
        Joint Acceleration & $\sum_i^{n_j} \ddot{q}_i$ & -2.5e-6 \\
        Action Rate & $\sum_i^{n_j} ||a_i^t - a_i^{t-1}||^2$  & -0.3 \\
        Energy & $\sum_i^{n_j} ||\tau_i\dot{q}_i||$ & -5.e-3 \\
        Undesired Collisions & $\sum_i^{n_c} \mathbbold{1}_{\text{i,contact}}$ & -1.0 \\
        Foot Clearance & $\sum^{n_f}_{i=1}S(-\sigma .h_{i}) . ||v^{xy}_{i}||$ & -5.0 \\
        Foot velocity & $\sum_i^{n_f}bf(-1.75+||v_f||,\mu)$ & -30.0 \\
    \end{tabular}
    \vspace{-15pt}
    \label{tab:rewards}
\end{table}

\subsection{Symmetry Augmentation} \label{subsect:methodology_symmetry}
To learn a more general policy, we sample from many contact plan parameters (see Table \ref{tab:sampling}), most of which is done at the beginning of each episode. This can limit the number of unique permutations seen during training, and so to help the learning process, we perform some symmetry augmentation on the data. Following \cite{mittal_symmetry_2024}, we augment with symmetry around the X-axis, Y-axis and the diagonal XY-axis. In essence, we mirror the observations (and corresponding actions) around those axes, and concatenate them to the original observations.
This has both an aesthetic effect and a performance benefit, which we analyze in Section \ref{subsect:results_ablation}.

\subsection{Training setup}
We use IsaacLab \cite{nvidia_isaac_2025} to train the policy for 12k update steps, with 24-step roll-outs across 4096 environments, and 15 epochs and 8 mini-batches per update. We use separate networks for the actor and the critic of size $\text{dim}=[512,256,128]$ each, with Mish activations. We use a classifier-based critic \cite{farebrother_stop_2024} with 51 bins to get a more stable value loss. We train on a single RTX 4090 for $\sim15$ hours.

\begin{figure*}[t!]
    \centering
        \resizebox{0.99\textwidth}{!}
    {
        \begin{tikzpicture}
        \node[anchor=south west, inner sep=0] (image) at (0,0) {\includegraphics[width=0.98\textwidth]{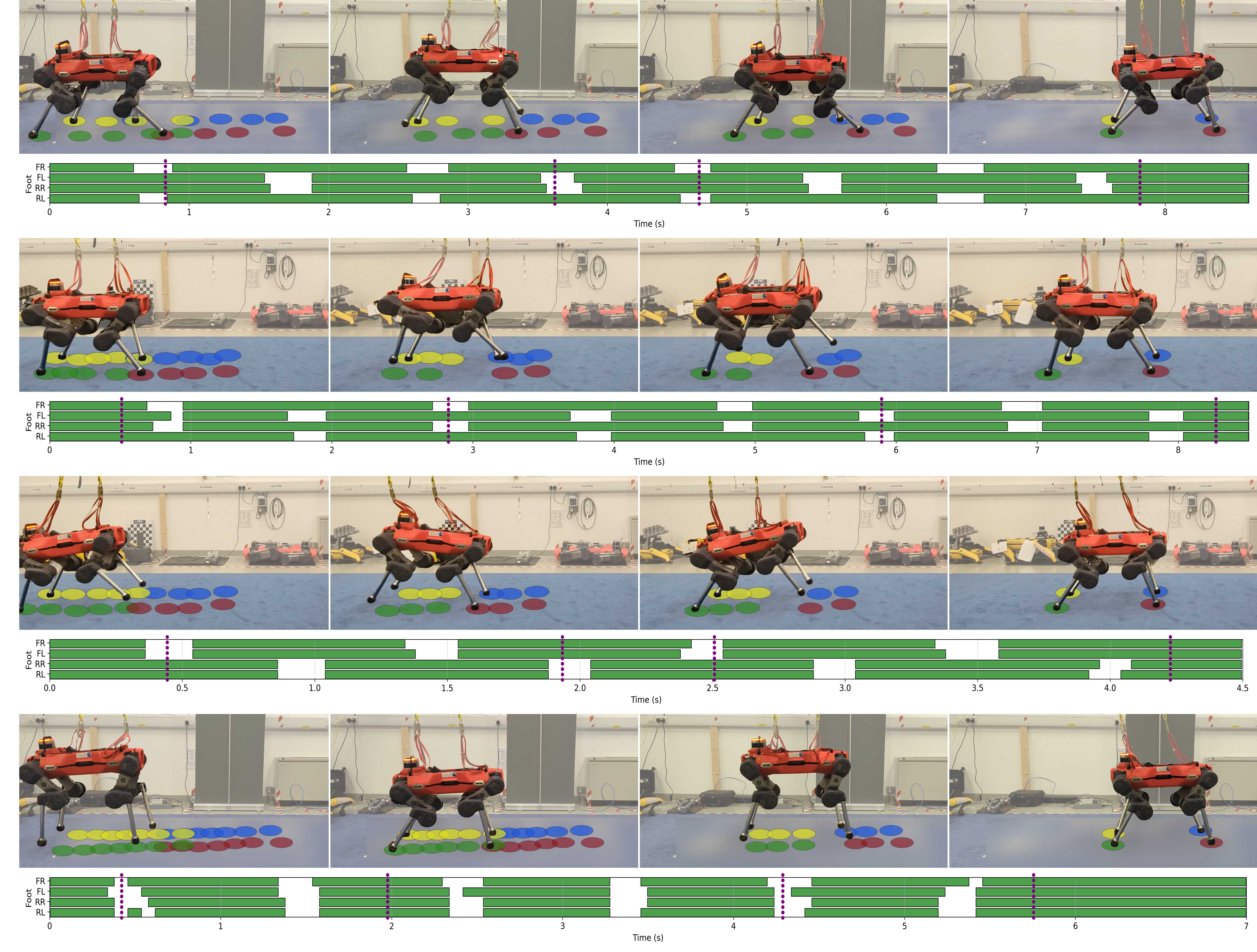}};

        \begin{scope}[x={(image.south east)},y={(image.north west)}]

            \node[
                rotate=90,          
                text width=1.8cm,   
                align=center        
            ] at (0.005, 0.919) {\rowlabel {Trot}};
            \node[
                rotate=90,          
                text width=2.cm,   
                align=center        
            ] at (0.005, 0.669) {\rowlabel{Pace}};
            \node[
                rotate=90,          
                text width=2.cm,   
                align=center        
            ] at (0.005, 0.419)  {\rowlabel{Bound}};
            \node[
                rotate=90,          
                text width=2.cm,   
                align=center        
            ] at (0.005, 0.169)  {\rowlabel{Pronk}};
        \end{scope}

    \end{tikzpicture}
    }
    \caption{GeCCo executing multiple different gaits, with their corresponding footfall pattern: \textit{Top row}: Trotting gait with alternating diagonal feet. \textit{Second row}: Pacing gait with alternating left/right side. \textit{Third row}: Bounding gait, with the front feet jumping first, followed by the rear feet. \textit{Bottom row}: Pronking/Jumping gait, with all feet taking off and touching down simultaneously. The circles illustrate the real contact locations.}
    \label{fig:results_gaits}
\end{figure*}
\section{Results}
\subsection{Experimental setup}
We deploy the policy on the ANYmal-D robot. \edit{As in sim, the initial footsteps are planned ahead of time, and then adjusted online depending on the terrain. At runtime, the stock proprietary state estimator is used to express desired contact locations in the base frame. When possible, we avoid replanning to isolate and stress-test the tracking policy without corrective feedback. GeCCo was able to track long-horizon plans lasting up to $\sim30$s without drift-induced failures.}
We compare against an end-to-end velocity-tracking baseline, trained as prescribed in IsaacLab \cite{nvidia_isaac_2025} for 12k update steps. We try two variants --- \textit{Baseline (Partial)} trained on the same terrains as GeCCo (slopes, stairs, rough terrain), and \textit{Baseline (Full)} which includes sparse terrains (stepping stones, beams).
\subsection{Locomotive gaits}
In Fig. \ref{fig:results_gaits}, we show ANYmal perform four different gaits -- namely trot, pace, bound, and pronk. Underneath each time-lapse, we show the footfall pattern, using the built-in state estimator. As can be seen, the robot can track different gaits, with little variation between synchronized feet. In the supplementary video, we show that GeCCo can also smoothly transition between the gaits.
\subsection{Complex terrains}
\begin{figure*}
    \centering
    \resizebox{0.99\textwidth}{!}
    {
    \begin{tikzpicture}
        \node[anchor=south west, inner sep=0] (image) at (0,0) {\includegraphics[width=\textwidth]{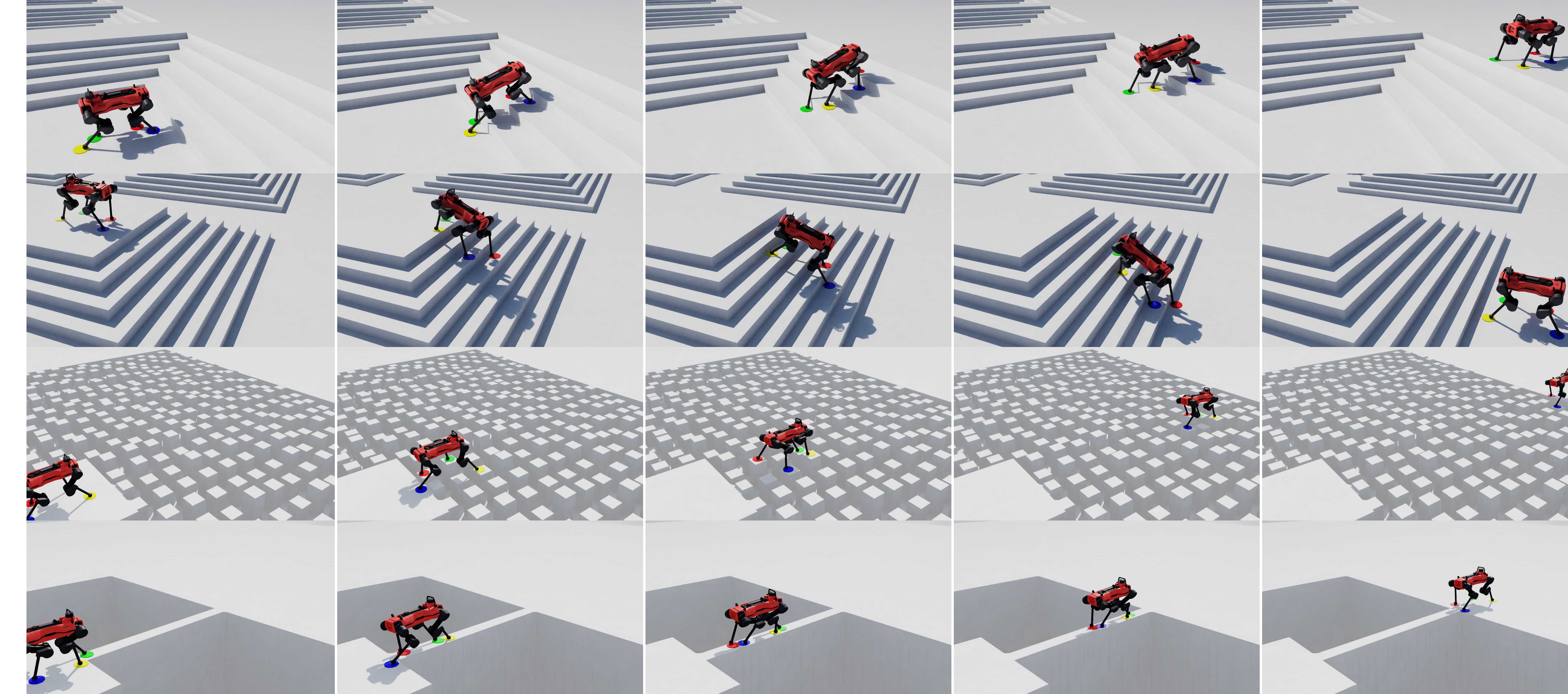}};

        \begin{scope}[x={(image.south east)},y={(image.north west)}]
            
            \node[
                rotate=90,          
                text width=1.25cm,   
                align=center        
            ] at (0.01, 0.875) {\small \rowlabel{Stairs Up}};
            \node[
                rotate=90,          
                text width=2cm,   
                align=center        
            ] at (0.01, 0.625)  {\small \rowlabel{Stairs Down}};
            \node[
                rotate=90,          
                text width=2.2cm,   
                align=center        
            ] at (0.01, 0.375) {\small \rowlabel{Step-Stones}};
            \node[
                rotate=90,          
                text width=2.2cm,   
                align=center        
            ] at (0.01, 0.125) {\small \rowlabel{Narrow Beam}};
        \end{scope}        
    \end{tikzpicture}
    }
    \caption{Traversing various complex terrains using GeCCo: \textit{Top row}: Climbing up stairs; \textit{Second row}: Climbing down a staircase; \textit{Third row}: Traversing unseen stepping stones; \textit{Bottom row}: Walking across an unseen narrow beam.}
    \label{fig:results_stairs}
\end{figure*}

To show the versatility of GeCCo, we evaluated its performance on a variety of complex terrains, such as the ones in the IsaacLab suite \cite{nvidia_isaac_2025}, as shown in Fig. \ref{fig:results_stairs}.
%
%
While sparse terrains are notoriously difficult for RL controllers \cite{zhang_learning_2023},  
GeCCo can easily traverse these terrains, as well. We tested two variations --- stepping stones and a narrow beam, neither of which was seen during training.
Nevertheless, as shown in Fig. \ref{fig:results_stairs}
, the policy can successfully traverse them in zero-shot, without any fine-tuning.

\begin{figure}
    \centering
    \vspace{4pt}
    \includegraphics[width=0.99\columnwidth]{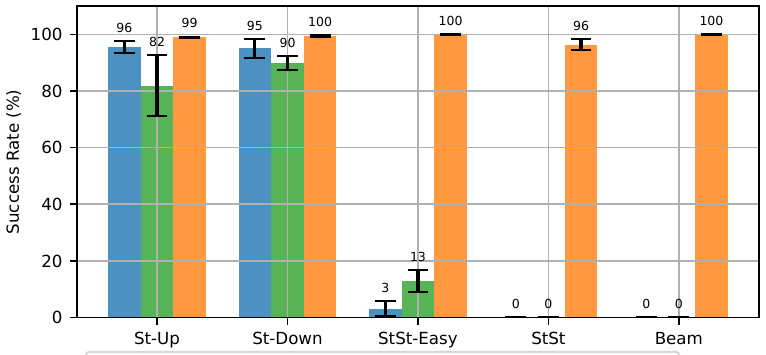}
    \caption{Results from 3x500 trials in simulation across five difficult terrains, showing the mean success rate and the $\pm$std. Success counts as walking 5\unit{m} without terminating.}
    \vspace{-4pt}
    \label{fig:results_terrains}
\end{figure}
In Fig. \ref{fig:results_terrains}, we report the quantitative results across five different terrains (similarly to those seen in Fig. \ref{fig:results_stairs}, with \textit{Stairs-Up} and \textit{Stairs-Down} having step width of 0.25\unit{m} and height of 0.2\unit{m}) for 3$\times$500 trials. We keep the planner fixed to a default stepping gait of length $\in [0.15,0.2]$\unit{m} and width of $0.2$\unit{m}. GeCCo achieves better results than an end-to-end velocity-tracking policy on the staircases, even with a simple footstep planner. On the more difficult, sparse terrains that have not been seen during training (\textit{Stepping-Stones-Easy} (0.8\unit{m} width stones, 0.05\unit{m} gap), \textit{Stepping-Stones} (0.35\unit{m} width stones, 0.2\unit{m} gap) and \textit{Narrow Beam} (0.30\unit{m} width)), GeCCo manages to maintain a similar success rate, while the Baseline (Partial) policy fails entirely. Baseline (Full) achieves $\sim18\%$ success on \textit{Stepping-Stones-Easy}, but still fails to traverse the more difficult ones. Prior work \cite{zhang_learning_2023} has shown that end-to-end policies require additional curricula, exploration terms, and fine-tuning to successfully learn to traverse such environments. Due to a lack of open-source implementation and specific details on terrain parameters, we could not directly compare to \cite{jenelten_dtc_2023,zhang_learning_2023, he_attention-based_2025}.
%
%
%

\begin{figure*}
    \centering
    \resizebox{0.987\textwidth}{!}
    {
    \begin{tikzpicture}
        \node[anchor=south west, inner sep=0] (image) at (0,0) {\includegraphics[width=0.99\textwidth]{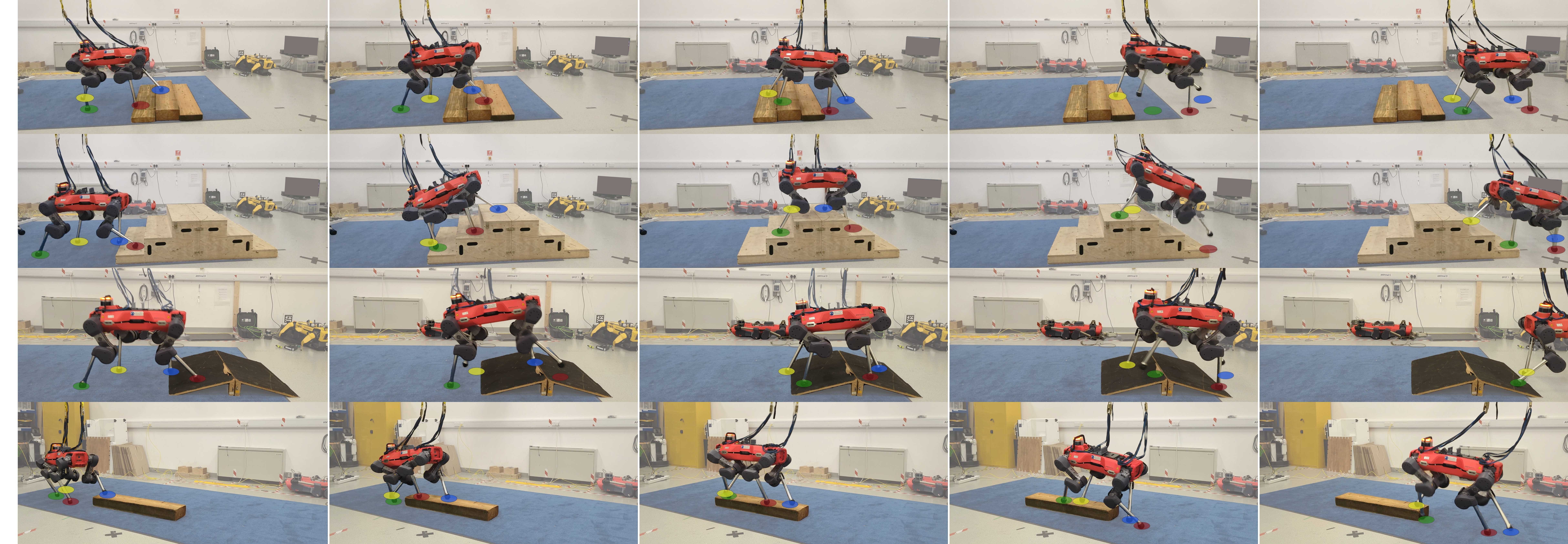}};

        \begin{scope}[x={(image.south east)},y={(image.north west)}]
            
            \node[
                rotate=90,          
                text width=1.08cm,   
                align=center        
            ] at (0.0, 0.875) {\small \rowlabel{Stairs-A}};
            \node[
                rotate=90,          
                text width=2cm,   
                align=center        
            ] at (0.0, 0.625)  {\small \rowlabel{Stairs-B}};
            \node[
                rotate=90,          
                text width=2.2cm,   
                align=center        
            ] at (0.0, 0.375) {\small \rowlabel{Ramp}};
            \node[
                rotate=90,          
                text width=2.2cm,   
                align=center        
            ] at (0.0, 0.125) {\small \rowlabel{Narrow Beam}};
        \end{scope}

    \end{tikzpicture}
    }
    \caption{Traversing various complex terrains using GeCCo: \textit{Top row}: Climbing up small stairs; \textit{Second row}: Climbing up a large staircase; \textit{Third row}: Walking up and down a ramp; \textit{Bottom row}: Walking across a narrow beam.}
    \label{fig:results_hardware_terrains}
\end{figure*}

In Fig. \ref{fig:results_hardware_terrains}, we demonstrate the successful hardware deployment on these much more difficult terrains. We use a simple footstep planner, which maintains a predefined gait, and uses the existing elevation map to adjust steps to the correct terrain height and away from edges. Despite the planner not considering any kino-dynamic constraints, our results show that GeCCo can track such plans even in adversarial terrains.
\subsection{Loco-manipulation} \label{subsect:results_locomani}
%
\begin{figure*}[htb!]
    \centering
    \resizebox{0.987\textwidth}{!}
    {
    \begin{tikzpicture}
        \node[anchor=south west, inner sep=0] (image) at (0,0) {\includegraphics[width=0.99\textwidth]{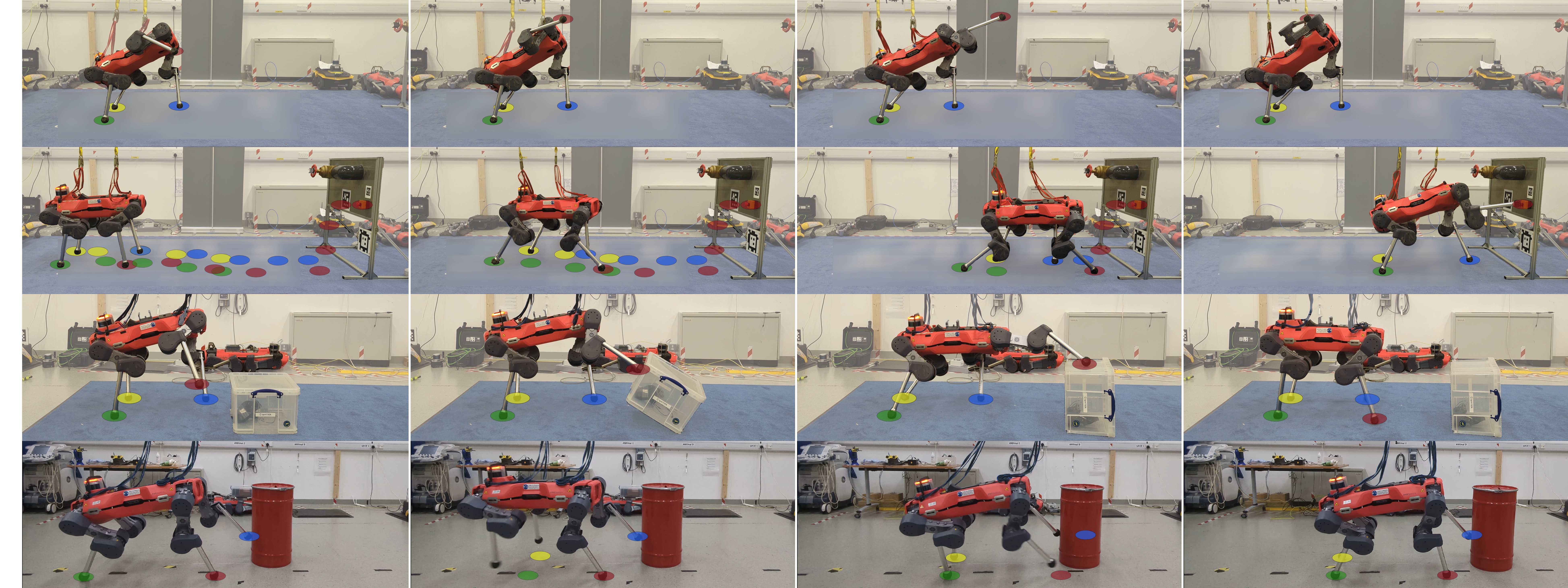}};

        \begin{scope}[x={(image.south east)},y={(image.north west)}]
            
            \node[
                rotate=90,          
                text width=1.25cm,   
                align=center        
            ] at (0.0, 0.875) {\small \rowlabel{Reach}};
            \node[
                rotate=90,          
                text width=2cm,   
                align=center        
            ] at (0.0, 0.625)  {\small \rowlabel{Button press}};
            \node[
                rotate=90,          
                text width=2.2cm,   
                align=center        
            ] at (0.0, 0.375) {\small \rowlabel{Box flip}};
            \node[
                rotate=90,          
                text width=2.2cm,   
                align=center        
            ] at (0.0, 0.125) {\small \rowlabel{Barrel push}};
            
        \end{scope}
    \end{tikzpicture}
    }
    \caption{Loco-manipulation experiments: \textit{Top row}: Reaching out for different locations in 3D space using the front right foot; \textit{Second row}: Button-pressing loco-manipulation task, consisting of walking to the board and then using its foot to press a button; \textit{Third row}: Flipping a box using the robot's foot; \textit{Bottom row}: Pushing a barrel continuously while the robot walks.}
    \vspace{-4.5pt}
    \label{fig:results_locomanipulation}
\end{figure*}
\edit{Our method can also seamlessly combine commands for locomotion and for manipulation together. In Fig. \ref{fig:results_locomanipulation}, we show four examples of loco-manipulation behaviors, such as reaching for different points in 3D space, walking to a push-button and pressing it using the foot, a non-prehensile task of flipping a box while standing in place, and pushing a barrel continuously while walking. The manipulation target for the push-button task is detected with an AprilTag, while the rest are either teleoperated or predefined. The barrel-push task shows the generalization capabilities of GeCCo --- despite the task being outside of the training distribution, the robot can adapt and push the object forward, while also balancing.}

%
%
%
\subsection{Ablation} \label{subsect:results_ablation}
\begin{figure}
    \centering
    \includegraphics[width=0.99\columnwidth]{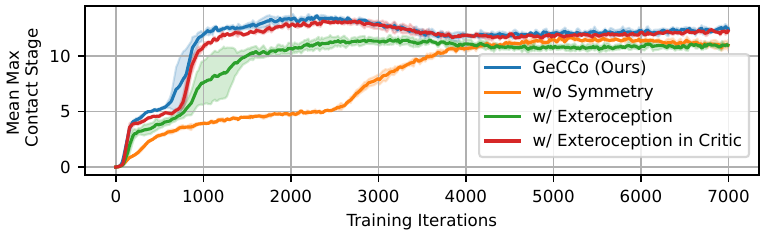}
    \caption{Ablation study showing the mean maximum contact stage achieved in an episode, with the shaded region showing the variance across 3 random seeds.}
    \vspace{-4.5pt}
    \label{fig:results_ablations}
\end{figure}

To understand what makes GeCCo work, we run several ablations (Fig. \ref{fig:results_ablations}), reporting the mean maximum contact stage per episode. In principle the contact plan abstracts the terrain away from the controller, but the robot might still benefit from extra information. \textbf{w/ Exteroception} adds height scans as an observation, yet performs slightly worse, likely due to the larger network input -- suggesting the contact plan already carries enough information to traverse complex terrains. It also limits generalization: the policy fails on the unseen stepping-stones and narrow-beam terrains, whose height scans are out of distribution. We also evaluated adding the Exteroception in the Critic only, which did not have an effect on the performance (\textbf{w/ Exteroception in Critic}). Ablation \textbf{w/o Symmetry} removes the symmetry augmentation (Section \ref{subsect:methodology_symmetry}). This degrades performance two-fold: aesthetically, and more importantly in contact tracking, likely due to differing training samples across feet -- confirming that symmetry augmentation aids more general policies.

\begin{figure}
    \centering
    \includegraphics[width=\columnwidth]{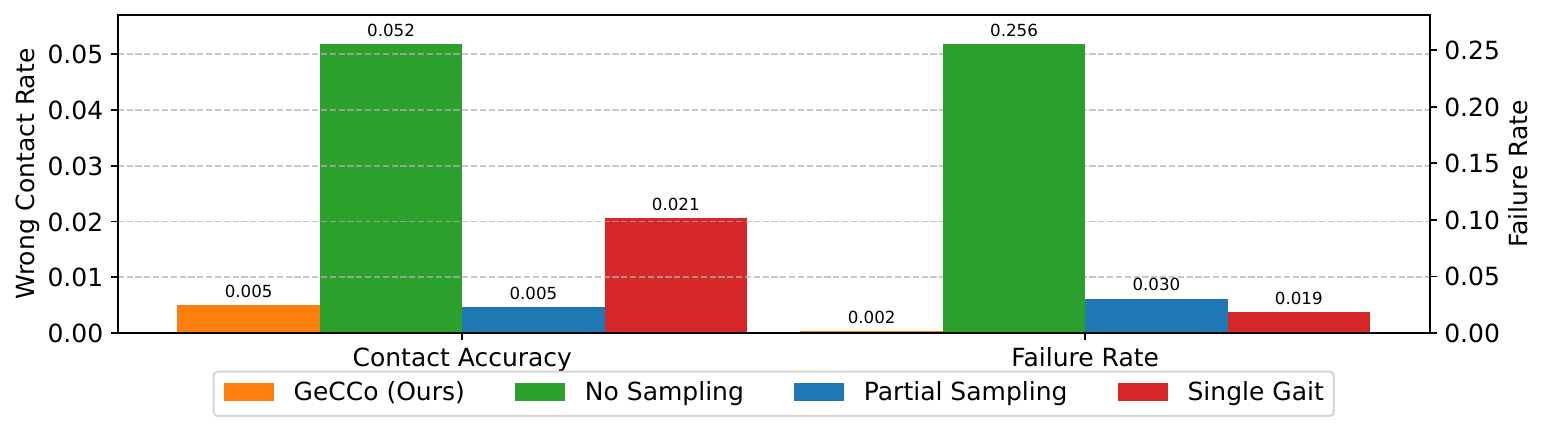}
    \caption{Ablation bar plots showing the contact tracking accuracy (left) -- in terms of median wrong contacts ratio, and the failure rate (right) across 10k envs.}
    \label{fig:ablation_sampling}
\end{figure}

\edit{We then ablate components of the sampling strategy (Table \ref{tab:sampling}) to isolate their contribution to robustness (Fig. \ref{fig:ablation_sampling}). We evaluate three variants: \textbf{Partial Sampling}, which fixes the base position/heading deviation and footstep width during training; \textbf{No Sampling}, which further removes footstep noise; and \textbf{Single Gait}, which only trains on a trotting gait. Across 10k environments we record the failure rate (falling) and the ratio of wrong footsteps. Without stochastic sampling the policy still performs reasonably well, as footstep plans are still adjusted based on the terrain, introducing contact variability. \textbf{Partial Sampling} matches GeCCo's tracking accuracy but is less robust, falling $\sim3\%$ of the time -- confirming that stochastic contact perturbations improve robustness. Sampling various gaits is crucial for generalizable tracking: the \textbf{Single Gait} baseline shows a higher failure rate and lower accuracy.}
\section{Conclusion}
In this work, we presented \text{GeCCo} -- a generalist contact-conditioned loco-manipulation policy trained with Deep RL. We demonstrated how this contact-based decomposition allows us to generalize to previously unseen terrains, like stepping-stones and narrow beams. With the same policy, our method can seamlessly switch between using its feet as non-prehensile manipulators and walking, capable of a variety of different gaits, like pronking, trotting, and pacing. Finally, our sampling-based planner enables doing all of this in a computationally efficient manner, allowing training within a few hours.

There are some limitations to GeCCo that we hope to address in the future. We only consider foot contacts - for more complex terrains the robot might need to use its shins or body to climb. Another limitation is that, while GeCCo can track arbitrary desired contact durations, the swing phase duration is uncontrolled. 
In future work, we would like to investigate how GeCCo can handle the notion of time better and allow for more precise control over swing phases. Finally, our next step would be to learn high-level planners for various loco-manipulation tasks on top of a pretrained GeCCo policy. As much of the complexity of locomotion has been shifted to this low-level tracker, one could train task-specific planners more easily. This is a direction we are particularly excited about, especially as a way to scale to more complex loco-manipulative interactions.

\section*{Acknowledgments}
We would like to thank Jacques Cloete, Aristotelis Papatheodorou, Reece O’Mahoney, and other members of the Oxford Robotics Institute for their helpful discussions and help with experiments.
\bibliographystyle{IEEEtran}

\begin{thebibliography}{10}
\providecommand{\url}[1]{#1}
\csname url@rmstyle\endcsname
\providecommand{\newblock}{\relax}
\providecommand{\bibinfo}[2]{#2}
\providecommand\BIBentrySTDinterwordspacing{\spaceskip=0pt\relax}
\providecommand\BIBentryALTinterwordstretchfactor{4}
\providecommand\BIBentryALTinterwordspacing{\spaceskip=\fontdimen2\font plus
\BIBentryALTinterwordstretchfactor\fontdimen3\font minus \fontdimen4\font\relax}
\providecommand\BIBforeignlanguage[2]{{%
\expandafter\ifx\csname l@#1\endcsname\relax
\typeout{** WARNING: IEEEtran.bst: No hyphenation pattern has been}%
\typeout{** loaded for the language `#1'. Using the pattern for}%
\typeout{** the default language instead.}%
\else
\language=\csname l@#1\endcsname
\fi
#2}}

\bibitem{rudin_cat-like_2022}
\BIBentryALTinterwordspacing
N.~Rudin, H.~Kolvenbach, V.~Tsounis, and M.~Hutter, ``Cat-{Like} {Jumping} and {Landing} of {Legged} {Robots} in {Low} {Gravity} {Using} {Deep} {Reinforcement} {Learning},'' \emph{IEEE Transactions on Robotics}, vol.~38, no.~1, pp. 317--328, 2022.
\BIBentrySTDinterwordspacing

\bibitem{atanassov_curriculum-based_2025}
\BIBentryALTinterwordspacing
V.~Atanassov, J.~Ding, J.~Kober, I.~Havoutis, and C.~D. Santina, ``Curriculum-{Based} {Reinforcement} {Learning} for {Quadrupedal} {Jumping}: {A} {Reference}-{Free} {Design},'' \emph{IEEE Robotics \& Automation Magazine}, vol.~32, no.~2, pp. 35--48, 2025.
\BIBentrySTDinterwordspacing

\bibitem{schwarke_curiosity-driven_2023}
\BIBentryALTinterwordspacing
C.~Schwarke, V.~Klemm, M.~v.~d. Boon, M.~Bjelonic, and M.~Hutter, ``\BIBforeignlanguage{en}{Curiosity-{Driven} {Learning} of {Joint} {Locomotion} and {Manipulation} {Tasks}},'' in \emph{\BIBforeignlanguage{en}{Proceedings of {The} 7th {Conference} on {Robot} {Learning}}}.\hskip 1em plus 0.5em minus 0.4em\relax PMLR, Dec. 2023, pp. 2594--2610.
\BIBentrySTDinterwordspacing

\bibitem{jeon_learning_2024}
\BIBentryALTinterwordspacing
S.~Jeon, M.~Jung, S.~Choi, B.~Kim, and J.~Hwangbo, ``Learning {Whole}-{Body} {Manipulation} for {Quadrupedal} {Robot},'' \emph{IEEE Robotics and Automation Letters}, vol.~9, no.~1, pp. 699--706, 2024.
\BIBentrySTDinterwordspacing

\bibitem{arm_pedipulate_2024}
\BIBentryALTinterwordspacing
P.~Arm, M.~Mittal, H.~Kolvenbach, and M.~Hutter, ``Pedipulate: {Enabling} {Manipulation} {Skills} using a {Quadruped} {Robot}’s {Leg},'' in \emph{2024 {IEEE} {International} {Conference} on {Robotics} and {Automation} ({ICRA})}.\hskip 1em plus 0.5em minus 0.4em\relax Yokohama, Japan: IEEE, May 2024, pp. 5717--5723.
\BIBentrySTDinterwordspacing

\bibitem{chen_learning_2020}
\BIBentryALTinterwordspacing
D.~Chen, B.~Zhou, V.~Koltun, and P.~Krähenbühl, ``\BIBforeignlanguage{en}{Learning by {Cheating}},'' in \emph{\BIBforeignlanguage{en}{Proceedings of the {Conference} on {Robot} {Learning}}}.\hskip 1em plus 0.5em minus 0.4em\relax PMLR, May 2020, pp. 66--75.
\BIBentrySTDinterwordspacing

\bibitem{miki_learning_2022}
\BIBentryALTinterwordspacing
T.~Miki, J.~Lee, J.~Hwangbo, L.~Wellhausen, V.~Koltun, and M.~Hutter, ``Learning robust perceptive locomotion for quadrupedal robots in the wild,'' \emph{Science Robotics}, vol.~7, no.~62, p. eabk2822, 2022.
\BIBentrySTDinterwordspacing

\bibitem{rudin_advanced_2022}
\BIBentryALTinterwordspacing
N.~Rudin, D.~Hoeller, M.~Bjelonic, and M.~Hutter, ``Advanced {Skills} by {Learning} {Locomotion} and {Local} {Navigation} {End}-to-{End},'' 2022, arXiv:2209.12827 [cs].
\BIBentrySTDinterwordspacing

\bibitem{atanassov_constrained_2024}
\BIBentryALTinterwordspacing
V.~Atanassov, W.~Yu, A.~L. Mitchell, M.~N. Finean, and I.~Havoutis, ``Constrained {Skill} {Discovery}: {Quadruped} {Locomotion} with {Unsupervised} {Reinforcement} {Learning},'' 2024, arXiv:2410.07877 [cs].
\BIBentrySTDinterwordspacing

\bibitem{hoeller_anymal_2023}
\BIBentryALTinterwordspacing
D.~Hoeller, N.~Rudin, D.~Sako, and M.~Hutter, ``{ANYmal} {Parkour}: {Learning} {Agile} {Navigation} for {Quadrupedal} {Robots},'' 2023, arXiv:2306.14874 [cs].
\BIBentrySTDinterwordspacing

\bibitem{jenelten_dtc_2023}
\BIBentryALTinterwordspacing
F.~Jenelten, J.~He, F.~Farshidian, and M.~Hutter, ``{DTC}: {Deep} {Tracking} {Control} -- {A} {Unifying} {Approach} to {Model}-{Based} {Planning} and {Reinforcement}-{Learning} for {Versatile} and {Robust} {Locomotion},'' 2023, arXiv:2309.15462 [cs, eess].
\BIBentrySTDinterwordspacing

\bibitem{zhang_wococo_2024}
\BIBentryALTinterwordspacing
C.~Zhang, W.~Xiao, T.~He, and G.~Shi, ``{WoCoCo}: {Learning} {Whole}-{Body} {Humanoid} {Control} with {Sequential} {Contacts},'' 2024.
\BIBentrySTDinterwordspacing

\bibitem{kim_high-speed_2025}
H.~Kim, H.~Oh, J.~Park, \emph{et~al.}, ``\BIBforeignlanguage{en}{High-speed control and navigation for quadrupedal robots on complex and discrete terrain},'' \emph{\BIBforeignlanguage{en}{Science RoboticS}}, 2025.

\bibitem{grandia_perceptive_2022}
\BIBentryALTinterwordspacing
R.~Grandia, F.~Jenelten, S.~Yang, F.~Farshidian, and M.~Hutter, ``Perceptive {Locomotion} through {Nonlinear} {Model} {Predictive} {Control},'' 2022, arXiv:2208.08373 [cs].
\BIBentrySTDinterwordspacing

\bibitem{papatheodorou_momentum-aware_2024}
\BIBentryALTinterwordspacing
A.~Papatheodorou, W.~Merkt, A.~L. Mitchell, and I.~Havoutis, ``Momentum-{Aware} {Trajectory} {Optimisation} using {Full}-{Centroidal} {Dynamics} and {Implicit} {Inverse} {Kinematics},'' in \emph{2024 {IEEE}/{RSJ} {International} {Conference} on {Intelligent} {Robots} and {Systems} ({IROS})}, Oct. 2024, pp. 11\,940--11\,947.
\BIBentrySTDinterwordspacing

\bibitem{hwangbo_learning_2019}
\BIBentryALTinterwordspacing
J.~Hwangbo, J.~Lee, A.~Dosovitskiy, \emph{et~al.}, ``Learning agile and dynamic motor skills for legged robots,'' \emph{Science Robotics}, vol.~4, no.~26, p. eaau5872, 2019.
\BIBentrySTDinterwordspacing

\bibitem{lee_learning_2020}
\BIBentryALTinterwordspacing
J.~Lee, J.~Hwangbo, L.~Wellhausen, V.~Koltun, and M.~Hutter, ``Learning quadrupedal locomotion over challenging terrain,'' \emph{Science Robotics}, vol.~5, no.~47, p. eabc5986, 2020.
\BIBentrySTDinterwordspacing

\bibitem{gangapurwala_rloc_2022}
S.~Gangapurwala, M.~Geisert, R.~Orsolino, M.~Fallon, and I.~Havoutis, ``{RLOC}: {Terrain}-{Aware} {Legged} {Locomotion} {Using} {Reinforcement} {Learning} and {Optimal} {Control},'' \emph{IEEE Transactions on Robotics}, vol.~38, no.~5, pp. 2908--2927, 2022.

\bibitem{zhang_learning_2023}
\BIBentryALTinterwordspacing
C.~Zhang, N.~Rudin, D.~Hoeller, and M.~Hutter, ``Learning {Agile} {Locomotion} on {Risky} {Terrains},'' 2023, arXiv:2311.10484 [cs].
\BIBentrySTDinterwordspacing

\bibitem{he_attention-based_2025}
\BIBentryALTinterwordspacing
J.~He, C.~Zhang, F.~Jenelten, R.~Grandia, M.~BÄcher, and M.~Hutter, ``Attention-{Based} {Map} {Encoding} for {Learning} {Generalized} {Legged} {Locomotion},'' 2025, arXiv:2506.09588 [cs].
\BIBentrySTDinterwordspacing

\bibitem{ji_dribblebot_2023}
\BIBentryALTinterwordspacing
Y.~Ji, G.~B. Margolis, and P.~Agrawal, ``{DribbleBot}: {Dynamic} {Legged} {Manipulation} in the {Wild},'' 2023, arXiv:2304.01159 [cs].
\BIBentrySTDinterwordspacing

\bibitem{cheng_legs_2023}
\BIBentryALTinterwordspacing
X.~Cheng, A.~Kumar, and D.~Pathak, ``Legs as {Manipulator}: {Pushing} {Quadrupedal} {Agility} {Beyond} {Locomotion},'' in \emph{2023 {IEEE} {International} {Conference} on {Robotics} and {Automation} ({ICRA})}.\hskip 1em plus 0.5em minus 0.4em\relax London, United Kingdom: IEEE, May 2023, pp. 5106--5112.
\BIBentrySTDinterwordspacing

\bibitem{cheng_rambo_2025}
\BIBentryALTinterwordspacing
J.~Cheng, D.~Kang, G.~Fadini, G.~Shi, and S.~Coros, ``{RAMBO}: {RL}-augmented {Model}-based {Whole}-body {Control} for {Loco}-manipulation,'' 2025, arXiv:2504.06662 [cs].
\BIBentrySTDinterwordspacing

\bibitem{pinto_asymmetric_2018}
\BIBentryALTinterwordspacing
L.~Pinto, M.~Andrychowicz, P.~Welinder, W.~Zaremba, and P.~Abbeel, ``Asymmetric {Actor} {Critic} for {Image}-{Based} {Robot} {Learning},'' in \emph{Robotics: {Science} and {Systems} {XIV}}.\hskip 1em plus 0.5em minus 0.4em\relax Robotics: Science and Systems Foundation, June 2018.
\BIBentrySTDinterwordspacing

\bibitem{rudin_learning_2022}
\BIBentryALTinterwordspacing
N.~Rudin, D.~Hoeller, P.~Reist, and M.~Hutter, ``\BIBforeignlanguage{en}{Learning to {Walk} in {Minutes} {Using} {Massively} {Parallel} {Deep} {Reinforcement} {Learning}},'' in \emph{\BIBforeignlanguage{en}{Proceedings of the 5th {Conference} on {Robot} {Learning}}}.\hskip 1em plus 0.5em minus 0.4em\relax PMLR, Jan. 2022, pp. 91--100.
\BIBentrySTDinterwordspacing

\bibitem{xue_full-order_2024}
\BIBentryALTinterwordspacing
H.~Xue, C.~Pan, Z.~Yi, G.~Qu, and G.~Shi, ``Full-{Order} {Sampling}-{Based} {MPC} for {Torque}-{Level} {Locomotion} {Control} via {Diffusion}-{Style} {Annealing},'' 2024, arXiv:2409.15610 [cs].
\BIBentrySTDinterwordspacing

\bibitem{kim_contact-implicit_2024}
\BIBentryALTinterwordspacing
G.~Kim, D.~Kang, J.-H. Kim, S.~Hong, and H.-W. Park, ``\BIBforeignlanguage{EN}{Contact-implicit {Model} {Predictive} {Control}: {Controlling} diverse quadruped motions without pre-planned contact modes or trajectories},'' \emph{\BIBforeignlanguage{EN}{The International Journal of Robotics Research}}, p. 02783649241273645, 2024.
\BIBentrySTDinterwordspacing

\bibitem{zhang_constrained_2024}
\BIBentryALTinterwordspacing
B.~Zhang, Y.~Zhang, L.~Frison, T.~Brox, and J.~Bödecker, ``Constrained {Reinforcement} {Learning} with {Smoothed} {Log} {Barrier} {Function},'' 2024, arXiv:2403.14508 [cs].
\BIBentrySTDinterwordspacing

\bibitem{mittal_symmetry_2024}
\BIBentryALTinterwordspacing
M.~Mittal, N.~Rudin, V.~Klemm, A.~Allshire, and M.~Hutter, ``Symmetry {Considerations} for {Learning} {Task} {Symmetric} {Robot} {Policies},'' 2024, arXiv:2403.04359.
\BIBentrySTDinterwordspacing

\bibitem{nvidia_isaac_2025}
\BIBentryALTinterwordspacing
NVIDIA, M.~Mittal, P.~Roth, \emph{et~al.}, ``Isaac {Lab}: {A} {GPU}-{Accelerated} {Simulation} {Framework} for {Multi}-{Modal} {Robot} {Learning},'' 2025, arXiv:2511.04831 [cs].
\BIBentrySTDinterwordspacing

\bibitem{farebrother_stop_2024}
\BIBentryALTinterwordspacing
J.~Farebrother, J.~Orbay, Q.~Vuong, \emph{et~al.}, ``Stop {Regressing}: {Training} {Value} {Functions} via {Classification} for {Scalable} {Deep} {RL},'' in \emph{Forty-first {International} {Conference} on {Machine} {Learning}, {ICML} 2024, {Vienna}, {Austria}, {July} 21-27, 2024}.\hskip 1em plus 0.5em minus 0.4em\relax OpenReview.net, 2024.
\BIBentrySTDinterwordspacing

\end{thebibliography}

\end{document}